\documentclass[letterpaper, 10 pt, conference]{ieeeconf}  
\usepackage{color-edits}
\usepackage[dvipsnames]{xcolor}
\definecolor{cmured1}{HTML}{C41230}
\definecolor{cmured2}{HTML}{941120}

\title{\LARGE \bf
Learning Multi-Agent Loco-Manipulation for Long-Horizon Quadrupedal Pushing
}

\author{Yuming Feng$^{1,*}$, Chuye Hong$^{1,*}$, Yaru Niu$^{1,*}$, Shiqi Liu$^{1}$, Yuxiang Yang$^{2}$, \\
Wenhao Yu$^{2}$, Tingnan Zhang$^{2}$, Jie Tan$^{2}$, and Ding Zhao$^{1}$ \\
$^{1}$Carnegie Mellon University, $^{2}$Google DeepMind, $^*$Equal contributions \\
\href{https://collaborative-mapush.github.io/}{\textcolor{cmured2}{https://collaborative-mapush.github.io}}
}

\usepackage{multirow}
\usepackage{booktabs}
\usepackage{subcaption}
\usepackage{pifont}
\usepackage{hyperref}
\usepackage[dvipsnames]{xcolor}
\usepackage{graphicx}
\usepackage{float}
\usepackage{amsmath}
\usepackage{amsfonts}
\usepackage{amssymb}
\usepackage{bbm}
\usepackage{cite}
\usepackage{caption}

\begin{document}

\thispagestyle{empty}
\pagestyle{empty}



\twocolumn[{
\renewcommand\twocolumn[1][]{#1}
\maketitle
\vspace{-0.22in}
\begin{center}
\noindent\begin{minipage}{1.0\textwidth}
    \includegraphics[width= 1.0\linewidth]{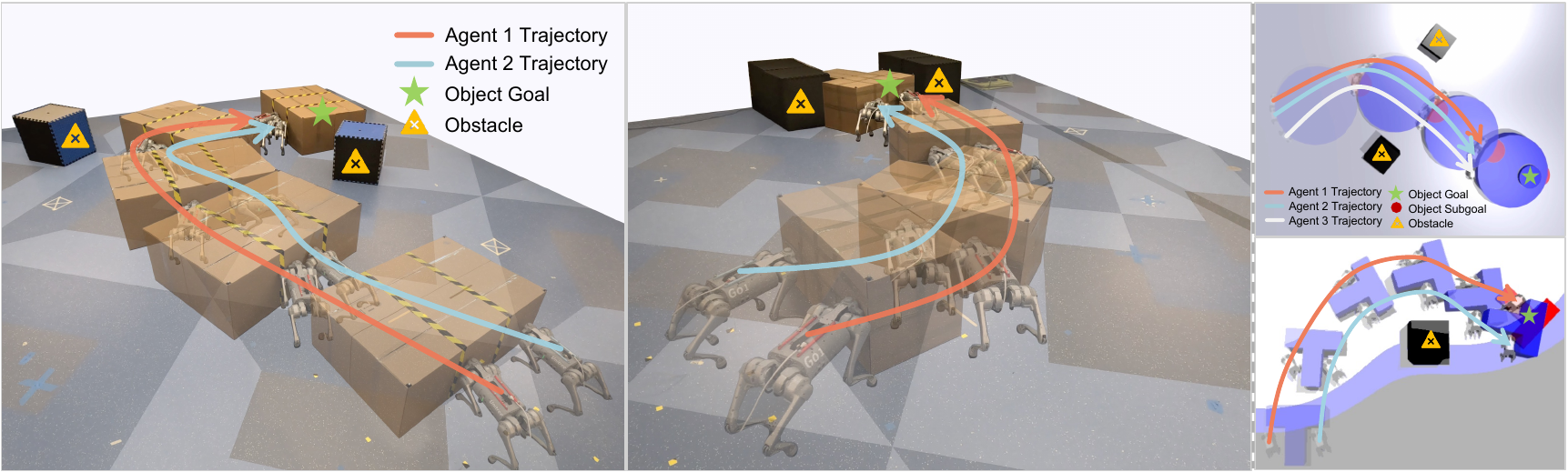}
    \end{minipage}\hspace{0.05in}
    \captionof{figure}{Our proposed method enables long-horizon collaborative pushing by multiple quadrupedal robots in environments with obstacles. The high-level controller within our hierarchical MARL framework generates adaptive subgoals to guide the lower-level policies during the collaborative manipulation of large objects of varying shapes. The agents' adaptive coordination ensures smooth obstacle avoidance and successful task completion, showcasing the robustness and flexibility of our hierarchical framework.}
    \label{fig:head}
    \vspace{-0.0in}
\end{center}
}]


\begin{abstract}
Recently, quadrupedal locomotion has achieved significant success, but their manipulation capabilities, particularly in handling large objects, remain limited, restricting their usefulness in demanding real-world applications such as search and rescue, construction, industrial automation, and room organization. This paper tackles the task of obstacle-aware, long-horizon pushing by multiple quadrupedal robots. 
We propose a hierarchical multi-agent reinforcement learning framework with three levels of control. The high-level controller integrates an RRT planner and a centralized adaptive policy to generate subgoals, while the mid-level controller uses a decentralized goal-conditioned policy to guide the robots toward these sub-goals. A pre-trained low-level locomotion policy executes the movement commands. We evaluate our method against several baselines in simulation, demonstrating significant improvements over baseline approaches, with $36.0\%$ higher success rates and $24.5\%$ reduction in completion time than the best baseline. Our framework successfully enables long-horizon, obstacle-aware manipulation tasks like Push-Cuboid and Push-T on Go1 robots in the real world.

\end{abstract}


\section{Introduction}
\label{sec:introduction}

Recent advances in quadrupedal robots have significantly improved their ability to traverse challenging terrains \cite{choi2023learning, lee2020learning, yang2023neural, kumar2021rma, lindqvist2022multimodality, cheng2024extreme}. While many studies have focused on enhancing their mobility and stability of locomotion, the manipulation capabilities of these robots remain relatively limited.
Efforts have been made to improve the quadrupedal capabilities in prehensile manipulation through attaching grippers or robotic arms on the robot \cite{bellicoso2019alma, mittal2022articulated, lin2024locoman, ha2024umi, fu2023deep, liu2024visual}, and non-prehensile manipulation by using legs \cite{shi2021circus, ji2023dribblebot, stolle2024perceptive, cheng2023legs} or the head \cite{rigo2023contact, sombolestan2023hierarchical} as the end-effectors. 
Although these advancements enable quadrupeds to handle some routine tasks, their limited ability to manipulate large and heavy objects still restricts their usefulness in demanding fields like search and rescue, construction, industrial automation, and room organization, where both dexterity and strength are essential. To address these challenges, researchers have explored adding support structures to the robots \cite{polverini2020multi, wolfslag2020optimisation}, coordinating whole-body movements \cite{jeon2023learning}, and using multiple robots \cite{sombolestan2023hierarchicalmulti, nachum2019multi} to strengthen contact forces and expand operational dimensions. However, achieving long-horizon manipulation of large objects in cluttered environments remains a largely unexplored and challenging task for quadrupeds.



In this work, we focus on addressing the challenge of obstacle-aware, long-horizon pushing by coordinating the whole-body motions of multiple quadrupedal robots. We build our work upon recent works of quadrupedal pushing that demonstrate impressive results. As shown in Table \ref{tab:comparisons}, while many approaches utilize multiple robots to enhance manipulation abilities, few focus on long-horizon pushing and obstacle avoidance, both of which are critical for real-world tasks. Additionally, the limited use of whole-body motions (e.g., relying solely on heads to push) \cite{sombolestan2023hierarchical, sombolestan2023hierarchicalmulti, nachum2019multi} restricts the contact patterns between robots and objects, making it difficult for the robots to perform diverse movements and avoid collisions with obstacles.
\begin{table}[htp]
\setlength{\tabcolsep}{3pt}
\centering
\caption{Comparisons between our proposed method and previous methods of quadrupedal pushing.}
\begin{tabular}{ccccc}
\toprule
\multirow{2}{*}{Method} & \multirow{2}{*}{Collaborative} & \multirow{2}{*}{\begin{tabular}[c]{@{}c@{}} Long- \\ Horizon \end{tabular}} & \multirow{2}{*}{\begin{tabular}[c]{@{}c@{}} Whole- \\ Body \end{tabular}} & \multirow{2}{*}{\begin{tabular}[c]{@{}c@{}} Obstacle- \\ Avoidance \end{tabular}}  \\ 
&&&& \\
\midrule
Sombolestan et al. \cite{sombolestan2023hierarchical} & \ding{55} & \ding{55} & \ding{55} & \ding{55} \\
Jeon et al. \cite{jeon2023learning} & \ding{55} & \ding{55} & \ding{51} & \ding{55} \\
Sombolestan et al. \cite{sombolestan2023hierarchicalmulti} & \ding{51} & \ding{55} & \ding{55} & \ding{55} \\
Nachum et al. \cite{nachum2019multi} & \ding{51} & \ding{51} & \ding{55} & \ding{51} \\
An et al. \cite{an2024solving} & \ding{51} & \ding{55} & \ding{51} & \ding{55} \\
Xiong et al. \cite{xiong2024mqe} & \ding{51} & \ding{55} & \ding{51} & \ding{55} \\
\textbf{Ours} & \ding{51} & \ding{51} & \ding{51} & \ding{51} \\
\bottomrule
\end{tabular}
\label{tab:comparisons}
\end{table}

To achieve collaborative, obstacle-aware, long-horizon quadrupedal pushing through whole-body motions, we propose a hierarchical multi-agent reinforcement learning (MARL) framework with three levels of controllers. The high-level controller integrates an Rapidly-exploring Random Tree (RRT) planner \cite{lavalle1998rapidly} and a centralized adaptive policy, which processes the reference trajectory, environment, and agent information to generate subgoals for the object. The mid-level controller learns a shared decentralized goal-conditioned policy, enabling multiple robots to coordinate and push the object toward the sequential subgoals proposed by the high-level controller. The low-level controller is a pre-trained locomotion policy that executes commands from the mid-level controller. We validate our approach through a series of experiments in both simulation and real-world tests on Go1 robots, a few of which are visualized in Figure \ref{fig:head}. Our results demonstrate that the proposed method achieves a $36.0\%$ higher success rate and a $24.5\%$ reduction in completion time compared to the best baseline approach in simulation. Furthermore, our method can be deployed on real robots to successfully complete obstacle-aware, long-horizon Push-Cuboid and Push-T tasks. The main contributions of this paper can be summarized as follows.

\begin{itemize}
    \item We propose a hierarchical MARL framework with three hierarchies that can handle long-horizon collaborative quadrupedal pushing in an environment with obstacles. 
    \item We benchmark our proposed method against baselines on various long-horizon pushing tasks involving obstacles in IsaacGym \cite{makoviychuk2021isaac}, demonstrating that our method significantly outperforms the baselines.
    \item We deploy our trained hierarchical policy on real robots, successfully completing the collaborative long-horizon Push-Cuboid and Push-T tasks with coordinated whole-body motions.
\end{itemize}

\begin{figure*}[htp]
    \centering
    \includegraphics[width=0.92\linewidth]{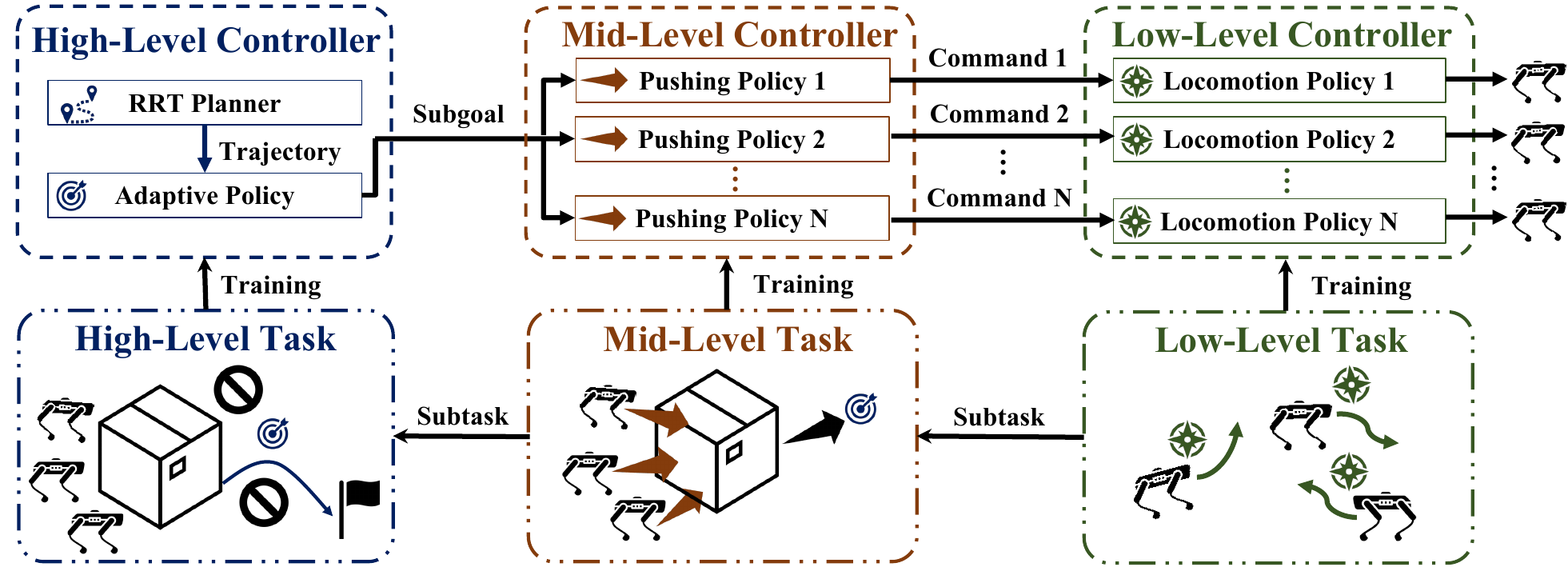}
    \caption{Overview of the proposed hierarchical MARL framework for collaborative long-horizon pushing tasks by quadrupedal robots. The framework comprises three layers: a \textbf{\textcolor{BlueViolet}{high-level controller}}, a \textbf{\textcolor{Sepia}{mid-level controller}}, and a \textbf{\textcolor{OliveGreen}{low-level controller}}. The \textbf{\textcolor{BlueViolet}{high-level controller}} utilizes an RRT planner to generate a trajectory and an adaptive policy to assign subgoals based on the dynamic states of the environment, object, and robots. The \textbf{\textcolor{Sepia}{mid-level controller}} employs decentralized pushing policies to convert a common subgoal into agent-specific velocity commands, which are then executed by the \textbf{\textcolor{OliveGreen}{low-level locomotion policy}} on each robot. Each layer is trained independently, leveraging frozen lower-level policies.
    }
    \label{fig:framework}
\end{figure*}

\section{Related Work}
\label{sec:related_work}
	\subsection{Loco-Manipulation for Legged Robots}
Researchers have proposed various optimization-based methods for prehensile loco-manipulation \cite{bellicoso2019alma, ferrolho2023roloma, mittal2022articulated, lin2024locoman}.
These approaches often use hierarchical structure to coordinate locomotion and gripper motions \cite{bellicoso2019alma}, decompose tracking objectives \cite{lin2024locoman}, or abstract object information for planning \cite{mittal2022articulated}.
Optimization-based methods have also been applied to single-robot non-prehensile manipulation tasks \cite{murooka2015whole, polverini2020multi, rigo2023contact, sombolestan2023hierarchical, wolfslag2020optimisation}, many of which rely on modeling and optimizing contacts with either the object or the ground. Murooka et al. demonstrate how humanoid robots can push large, heavy objects through contact posture planning \cite{murooka2015whole}, while Polverini et al. introduce a multi-contact controller for a centaur-type humanoid robot to handle similar tasks \cite{polverini2020multi}. 
Rigo et al. introduce a hierarchical MPC framework for optimizing contact in quadrupedal loco-manipulation, where the robot is constrained to using its head for pushing \cite{rigo2023contact}.

Recently, learning-based methods have demonstrated its effectiveness in loco-manipulation for legged robots.
Specifically, reinforcement learning (RL) has been used to train short-horizon quadrupedal pushing skills \cite{kumar2023cascaded, huang2023bayrntune, stolle2024perceptive}, and other non-prehensile loco-manipulation skills such as dribbling a soccer ball \cite{ji2023dribblebot}, manipulating a yoga ball \cite{shi2021circus} pressing buttons \cite{cheng2023legs}, opening doors \cite{zhang2024learning}, and carrying boxes \cite{zhang2024wococo}. Jeon et al. propose a hierarchical reinforcement learning framework for quadrupedal whole-body manipulation of large objects, capable of inferring manipulation-relevant privileged information through interaction history \cite{jeon2023learning}. Moreover, learning-based whole-body controllers are trained for prehensile manipulation that requires grasping various objects \cite{fu2023deep,arm2024pedipulate} and consuming visual inputs \cite{liu2024visual, ha2024umi, zhang2024gamma}. Our work focuses on quadrupedal pushing, coordinating whole-body motions using RL-trained policies without explicitly modeling contacts.

\subsection{Multi-Agent Collaborative Manipulation}
Optimization-based methods have proven effective in multi-agent collaborative manipulation across various robotic embodiments, such as mobile robots \cite{culbertson2018decentralized, culbertson2021decentralized, chen2015occlusion, Tang-RSS-24}, robotic arms \cite{yan2021decentralized}, quadrotors \cite{shorinwa2021distributed}, and six-legged robots \cite{mataric1995cooperative}.
Some works explore utilizing Model Predictive Control (MPC) to achieve cooperative locomotion for multiple quadrupedal robots holonomically constrained to one another \cite{fawcett2023distributed, kim2022cooperative, kim2023layered} or collaborative loco-manipulation with objects rigidly attached to each robotic hand \cite{de2023centralized}. However, these approaches might lack generalizability to more typical scenarios due to their reliance on specific inter-robot connections. The work in \cite{sombolestan2023hierarchicalmulti} introduces a hierarchical adaptive control method enabling multiple quadrupeds to cooperatively push an object with unknown properties along a predetermined path, though the robots are constrained to use their head to push the objects, while our task requires more diverse contact patterns.

Moreover, MARL are employed in cooperative bimanual manipulation for robotic arms \cite{ding2020distributed, zhang2021dair, li2023efficient, lee2019learning} and dexterous hands \cite{huang2023dynamic, chen2022towards}, and collaborative loco-manipulation for quadrupeds \cite{xiong2024mqe, an2024solving, nachum2019multi, ji2021reinforcement}, snake robots \cite{zhang2024composer} and bipedal robots \cite{pandit2024learning}.
Nachum et al. propose a two-level hierarchical policy in which the high-level policy generates subgoals for each robot to navigate toward \cite{nachum2019multi}. Xiong et al. benchmark MARL with a two-level hierarchical structure in cooperative and competitive tasks, but the methods struggle in a simple box-pushing scenario \cite{xiong2024mqe}. An et al. introduce a permutation-invariant network architecture that enables short-horizon multi-object pushing for wheeled-legged quadrupeds \cite{an2024solving}. However, these approaches primarily focus on generating effective robot-centric commands for locomotion controllers, making them limited in longer-horizon manipulation tasks. Our approach addresses these limitations by enabling multiple robots to coordinate whole-body motions for long-horizon pushing tasks in the environments with obstacles.


\subsection{Hierarchical Reinforcement Learning}
Hierarchical reinforcement learning (HRL) is often utilized to tackle challenging long-horizon decision making problems. In HRL methods, the high-level policy usually learns to set subgoals for the low-level \cite{nachum2018data, nachum2018near, levy2017learning, vezhnevets2017feudal}, or learns to combine and chain behavior primitives \cite{nasiriany2022augmenting, lee2019learning, strudel2020learning}. In multi-agent settings, the high-level policy in hierarchical MARL generates goals or commands in a decentralized manner  \cite{an2024solving, xiong2024mqe, tang2018hierarchical}, or through a centralized controller \cite{ahilan2019feudal, ma2020feudal, nachum2019multi, liu2021coach}. Meanwhile, many learning-based controllers for legged robots follow a hierarchical structure, where a high-level RL policy provides intermediate commands to the low-level controllers, such as torso velocities \cite{jeon2023learning, an2024solving}, foot landing positions \cite{yu2021visual, gangapurwala2022rloc}, target poses \cite{tan2021hierarchical, fu2024humanplus}, gait timing \cite{yang2022fast, da2021learning} or a combination of several \cite{yang2023cajun, margolis2023walk}. 
In our approach, we use a centralized high-level controller to propose a shared object-centric goal for all robots, while decentralized mid-level controllers send torso velocity commands to each robot's low-level policy.






\section{Methodology}
\label{sec:methodology}
	
\subsection{Hierarchical Reinforcement Learning for Long-Horizon Multi-Robot Pushing}
To enable quadrupedal robots to collaboratively perform long-horizon pushing tasks in environments with obstacles, we propose a hierarchical reinforcement learning framework, as illustrated in Figure \ref{fig:framework}.  
This framework consists of three layers of controllers. At the top level, an RRT planner generates a geometrically feasible trajectory without accounting for the robots' pushing capabilities or the dynamics of multiple robots and the object. 
The high-level adaptive policy then uses this trajectory as a reference to assign a subgoal for the target object, based on the dynamic states of the environment, object, and robots.
Using this common subgoal, each robot's mid-level pushing policy provides velocity commands to its corresponding low-level policy.
Due to the computational demands of the RRT planner, it is executed only once at the start of each episode. Both the high-level adaptive policy and the mid-level controller operate at a frequency of 50 Hz, with the higher frequency at the high level proving beneficial for more adaptive behavior in our settings. The low-level locomotion policy also runs at 50 Hz, while the PD controller is implemented at 200 Hz in simulation and 2000 Hz on the physical robot.
In the following sections, we will introduce each of these three hierarchies in detail.

\subsection{Low-Level Controller}
\label{sec:low-level}

The low-level controller controls each robot individually to track the mid-level velocity commands.
More specifically, each low-level controller $\pi^{l,i}_{\varphi}: \mathcal{O}^{l,i}\rightarrow\mathcal{A}^{l,i}$ computes motor commands $a^{l,i}$ to track the mid-level velocity command $a^{m,i}=(v^i_{x}, v^i_{y}, v^i_{\text{yaw}})$.
Despite recent progress of learning-based low-level controllers \cite{margolis2023walk}, we find these controllers to suffer from a large sim-to-real gap, and cannot accurately track the velocity commands, especially when the robot is pushing a heavy object.
Instead, we use Unitree's built-in low-level controller, which tracks the velocity commands significantly more robustly in the real world.
For efficient policy training in simulation, we train a learned low-level policy to mimic the behavior of Unitree's built-in controller.

We create the simulated low-level controller based on \emph{Walk-These-Ways} (WTW) \cite{margolis2023walk}.
As an RL framework for low-level motor control, WTW can learn locomotion behaviors with configurable body pose, gait timing, and reference velocity.
We measure these parameters on the built-in controller, and reproduce them in WTW to learn similar behaviors.
During upper-level policy training, we invoke the low-level WTW policy in parallel on GPU, which significantly reduced the training time.

\subsection{Mid-Level Controller}
The mid-level controller $\pi^{m,i}_{\phi}: \mathcal{O}^{m,i}\rightarrow\mathcal{A}^{m,i}$ is a decentralized policy of agent $i$, where $\mathcal{O}^{m,i}$ represents the mid-level local observation space of robot $i$, and $A^{m,i}$ is the action space of the mid-level policy of agent $i$. This decentralized policy takes as input the high-level action $a^{h}$, the local observation of robot $i$, $o^{m,i}\in\mathcal{O}^{m,i}$, which consists of the local observations of the target object state $s^i_{\text{object}}$, obstacle state $s^i_{\text{obstacle}}$, and the state of other robots, $\{s^i_{j}\}^N_{j=1,j\neq i}$, all computed in the local torso frame of the robot $i$. For example, $s^i_{\text{object}}$ can be expanded as $(x^i_{\text{object}}, y^i_{\text{object}}, \psi^{i}_{\text{object}})$, where $r^i_{\text{object}}=(x^i_{\text{object}}, y^i_{\text{object}})$ is the 2D position of the object, and $\psi^i_{\text{object}}$ is its yaw angle, both in the local frame of robot $i$. The mid-level policy of agent $i$ will output a mid-level action $a^{m,i}\sim\pi^{m,i}_{\phi}(a^{m,i}|o^{m,i},a^h)\in\mathcal{A}^{m,i}$ in a decentralized manner
to the low-level controller of robot $i$. 

In practice, we train a mid-level policy shared by all robots, noted as $\pi^m_{\phi}$.
Following the scheme of centralized training and decentralized execution, it is trained by MAPPO \cite{yu2022surprising} to optimize the objective function $\mathcal{J}^m(\theta)=\mathbb{E}_{\tau\sim\rho_\pi}\left[\sum_{t=0}^T{\gamma^tr^m(s_t, a^h_{t}, \{a^{m,i}_t\}^N_{i=1})}\right]$, where $s_t$ is a joint state at time $t$, $\tau$ is the trajectory sampled from a distribution $\rho_\pi$ induced by policy $\pi^m_{\phi}$, initial state $\rho^m_0$ and the transition probability $p^m$ that are defined by the mid-level task. Here, $r^m(\cdot)$ represents the reward function for the mid-level controller. During training, we randomly sample the subgoals of the object as $a^h$ and freeze the low-level policy. Meanwhile, we specialize the domain randomization for frictions to reduce the Sim2Real gap of pushing.

\subsection{High-Level Controller}
The high-level controller is composed of two elements, a RRT planner $\mathcal{P}:\mathcal{M}\times\mathcal{G}\rightarrow \mathcal{T}$ and a centralized adaptive policy $\pi^{h}_{\theta}:\mathcal{M}\times\mathcal{T}\times \mathcal{S}_{\text{object}}\times \mathcal{S}_1\times \mathcal{S}_2\times \cdots \mathcal{S}_N\rightarrow \mathcal{A}^h$, where $\mathcal{M}$ represents the map information space, $\mathcal{G}$ represents the goal space of the target object, $\mathcal{T}$ represents the trajectory space of the RRT planner, $\mathcal{S}_{\text{object}}$ denotes the object state space, $\mathcal{S}_i$ is the state space of robot $i$, and $A^h$ is the action space of the high-level adaptive policy.

The RRT planner takes the desired goal position of the object $g_{\text{object}}\in\mathcal{G}$ and the map information $p_{\text{map}}\in\mathcal{M}$ encompassing the obstacle position and the initial position of the object as input and outputs a reference trajectory $\tau_r\in\mathcal{T}$ for the adaptive policy $\pi^h_{\theta}$. The adaptive policy will use the desired goal $g_{\text{object}}$, each robot state $s_i\in \mathcal{S}_i$, the map information $p_{\text{map}}$ and the dynamic global object pose $s_{\text{object}}\in \mathcal{S}_{\text{object}}$, as the input, and output a high-level action $a^h\sim \pi^h_{\theta}(a^h|g_{\text{object}}, p_{\text{map}}, s_1, s_2, \cdots, s_N, s_{\text{object}}, \tau_r)\in\mathcal{A}^h$ as the subgoal position of the target object to the mid-level policy $\pi^m_{\phi}$.
The high-level adaptive policy is a centralized policy and trained via PPO \cite{schulman2017proximal} to optimize the objective function $\mathcal{J}^h(\theta)=\mathbb{E}_{\tau\sim\rho_\pi}\left[\sum_{t=0}^T{\gamma^tr^h(g_{\text{object},t}, p_{\text{map},t}, s_{\text{object},t}, \{s_{i,t}\}^N_{i=1},a^h_t, \tau_r)}\right]$ , where $\tau$ is the trajectory sampled from a distribution $\rho_\pi$ induced by policy $\pi^h_{\theta}$, initial state $\rho^h_0$ and the transition probability $p^h$ that are defined by the high-level task, Here, $r^h(\cdot)$ represents the reward function for the high-level controller. During training, we freeze the mid-level and low-level controller.


\label{subsec:reward}

\begin{figure}[htp]
    \centering
    \includegraphics[width=0.95\linewidth]{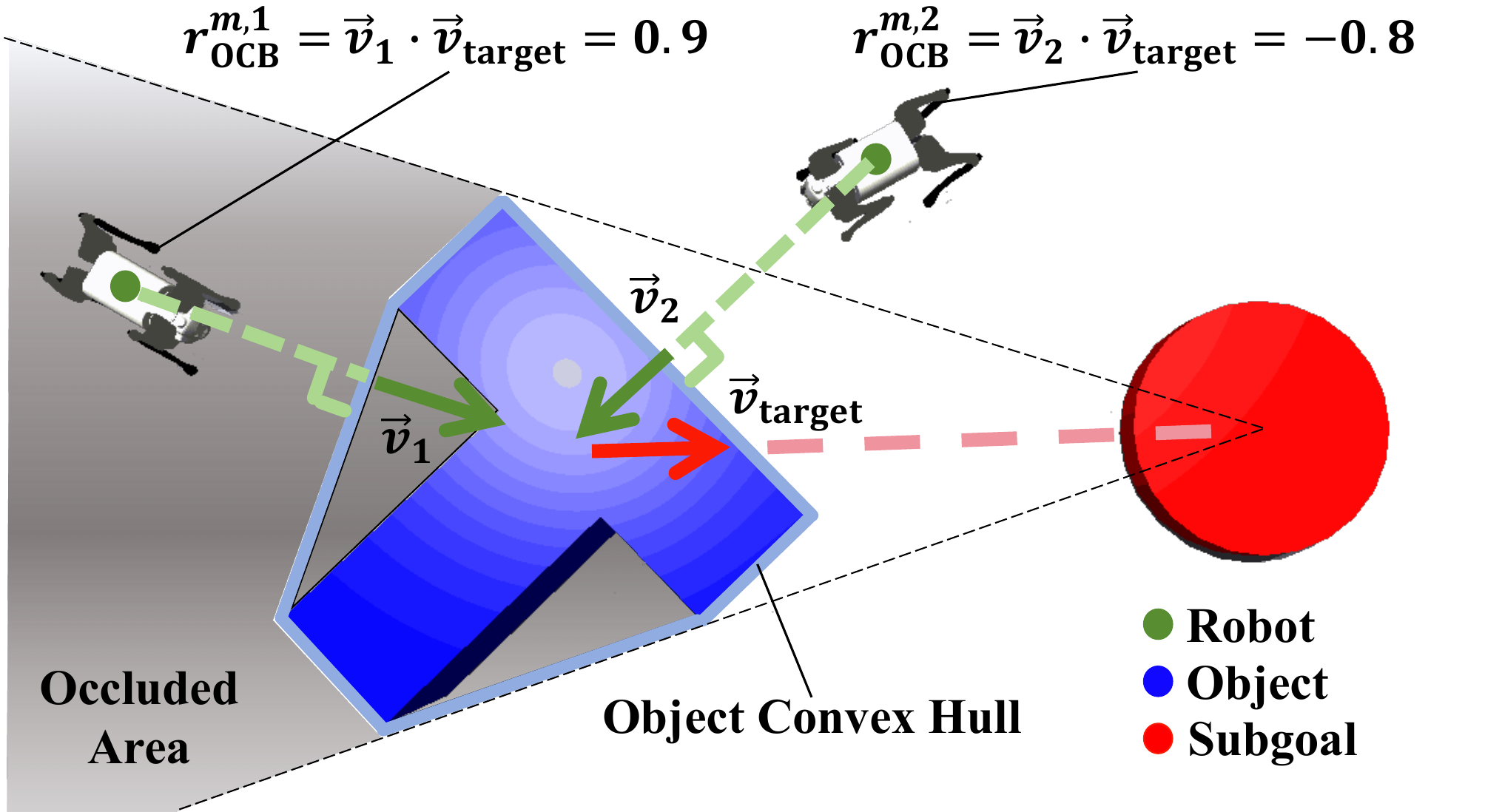}
    \caption{An example of the OCB reward. The robots are encouraged to push along object's convex hull perimeter that occludes their view of the subgoal, guiding the object's motion approximately in that direction. Here, $\vec{v}_{\text{target}}$ is a unit vector directing from the object towards the subgoal, , while $\vec{v}_i$ is a unit normal vector at the closest point on the object's convex hull to robot $i$, directed inward.}
    \label{fig:ocb}
\end{figure}
\subsection{Reward Design}
\subsubsection{Mid-Level Reward}
\label{subsubsec:mid-level reward}
Our mid-level reward function is formulated as $r^m = r^m_{\text{task}} + r^m_{\text{penalty}} + r^m_{\text{heuristic}}$. The mid-level task reward $r^m_{\text{task}}$ incentivizes actions that move the object toward and reach the target point, while the penalty term $r^m_{\text{penalty}}$ penalizes agents for close proximities, as well as for exceptions such as robot fall-overs and timeouts.

The mid-level heuristic reward \(r^m_{\text{heuristic}}\) plays a vital role in the pushing process, given the expansive action space and the inherent uncertainty and complexity of interactions during pushing. It is defined as \(r^m_{\text{heuristic}} = r^m_{\text{approach}} + r^m_{\text{vel}} + r^m_{\text{OCB}}\), where the mid-level approaching reward \(r^m_{\text{approach}}\) encourages agents to approach the object, and the velocity reward \(r^m_{\text{vel}}\) rewards agents when the object's velocity exceeds a predefined threshold, promoting diverse pushing actions while preventing oscillation near the object.

Importantly, an occlusion-based (OCB) reward \(r^m_{\text{OCB}}\), inspired by \cite{chen2015occlusion}, is introduced to guide agents toward more favorable contact points in the areas where robots' views of the subgoals are blocked. Specifically, the OCB reward of robot $i$ is calculated as $r^{m,i}_{\text{OCB}}=\vec{v}_i\cdot\vec{v}_{\text{target}}$, where $\vec{v}_i$ is the unit normal vector at the closet point on the object's convex hall to robot $i$ and $\vec{v}_{\text{target}}$ is unit vector directing from the object towards the subgoal, as depicted in Figure \ref{fig:ocb}.
Agents are rewarded or penalized based on \(r^m_{\text{OCB}}\), which is crucial in pushing tasks where identifying optimal contact points is challenging. This reward encourages agents to target occluded surfaces, enabling more effective pushing behavior.

\begin{table*}[htp]
\centering
\caption{Success rates and completion time ($\pm$ standard deviation) of our method and baselines in different settings in simulation. The completion time is scaled to $[0,1]$ where $1$ means taking up a full episode reaching the timeout.}
\label{tab:main_result}
\begin{tabular}{cc|cccccc}
\toprule
\multicolumn{2}{c|}{Task}                    & SA           & H+L             & M+L            & H+L FT             & M+L FT            & \textbf{Ours}           \\
\midrule
\multirow{2}{*}{Cuboid}     & S.R.$\uparrow$   & 4.5$\pm$0.3\%  & 0.23$\pm$0.02\% & 10.5$\pm$8.5\% & 41.0$\pm$16.2\% & 24.3$\pm$20.9\% & \textbf{77.5$\pm$3.0\%} \\
                          & C.T.$\downarrow$ & 0.98$\pm$0.02  & 1.00$\pm$0.00   & 0.95$\pm$0.04  & 0.76$\pm$0.12   & 0.86$\pm$0.12   & \textbf{0.66$\pm$0.04} \\
\midrule
\multirow{2}{*}{T-Shape}  & S.R.$\uparrow$   & 21.2$\pm$9.3\% & 9.0$\pm$6.0\%   & 1.8$\pm$1.0\%  & 7.3$\pm$3.8\%   & 25.8$\pm$28.9\% & \textbf{63.5$\pm$7.7\%} \\
                          & C.T.$\downarrow$ & 0.96$\pm$0.04  & 0.94$\pm$0.04   & 0.99$\pm$0.01  & 0.95$\pm$0.02   & 0.80$\pm$0.19   & \textbf{0.68$\pm$0.04} \\
\midrule
\multirow{2}{*}{Cylinder} & S.R.$\uparrow$   & 0.0$\pm$0.0\%  & 3.0$\pm$4.0\%   & 3.0$\pm$3.0\%  & 56.0$\pm$16.5\% & 26.9$\pm$27.3\%  & \textbf{71.2$\pm$5.1\%} \\
                          & C.T.$\downarrow$ & 1.00$\pm$0.00  & 0.99$\pm$0.02   & 0.80$\pm$0.02  & 0.70$\pm$0.14   & 0.81$\pm$0.20  & \textbf{0.48$\pm$0.01} \\ 
\bottomrule
\end{tabular}
\end{table*}

\subsubsection{High-Level Reward} 
\label{high-level reward}
\label{subsubsec:high-level reward}
The high-level reward function is composed of two terms: \(r^h = r^h_{\text{task}} + r^h_{\text{penalty}}\). The high-level task reward \(r^h_{\text{task}}\) provides a sparse reward for reaching the final target and two dense rewards: one for minimizing the distance between the subgoal and the nearest point on the RRT trajectory, and the other for reducing the distance between the object and the final target. This guides the robots to follow the RRT trajectory while allowing minor deviations for handling push complexities.
The high-level penalty \(r^h_{\text{penalty}}\) includes penalties for close distances to obstacles and severe punishments for exceptions such as robot fall-overs, collisions, object tilting, and timeouts.


\section{Experiments}
\label{sec:experiments}

\subsection{Simulation Setups}
\subsubsection{Environments and Tasks}\label{subsubsec:env and task}
We build our simulation environments in IsaacGym \cite{makoviychuk2021isaac}.
We consider a cluttered environment with randomly placed \(1.0\,\text{m} \times 1.0\,\text{m}\times 1.0\,\text{m}\) obstacles, where multiple quadrupeds need to push a target object to a desired goal. Unitree Go1 robots are utilized in simulation to match the physical robots, each with an approximate payload capacity of 5$\,$kg. 
The robots are tested with three types of objects varying
in shape and mass: a 4$\,$kg cuboid, a 3$\,$kg T-shaped block, and a 10$\,$kg cylinder with a radius of 1.5$\,$m. Each object is larger than the robot in size and close to or exceeds the robot's payload capacity.
Different numbers of agents are evaluated across tasks, with two agents for the cuboid and T-shaped block, and up-to four agents for the cylinder.
The initial positions and postures of the agents and target objects are randomly set within an area on one side of the room, while the target goals for the object are generated on the other side. The task is considered successful if the center of the object is positioned within 1$\,$m of the target. Failure occurs in the situation described in Sec. \ref{subsubsec:high-level reward}. The tasks are designed for challenging long-horizon pushing, with initial-to-target distances exceeding 10$\,$m. 


\subsubsection{Baselines}
We compare our proposed method with the following baselines.

\textbf{Single-Agent (SA)} retains the three hierarchical levels of the policy and the reward function design, but only a single quadrupedal robot is employed for each task.

\textbf{High-Level+Low-Level (H+L)} utilizes both a high-level and a low-level policy, where the high-level policy proposes subgoals for the robots, and the low-level policy aids the robots in navigating to these subgoals. We maintain $r^h_{\text{task}}$ and $r^h_{\text{penalty}}$ mentioned in Sec. \ref{subsubsec:high-level reward}. This baseline follows a similar approach to \cite{nachum2019multi}, with a multi-agent implementation using MAPPO \cite{yu2022surprising}.

\textbf{Mid-Level+Low-Level (M+L)} retains the mid-level and low-level policies without using a high-level policy to provide subgoals, meaning the robots are guided directly by the distant final target. 
Similar to the methods proposed in \cite{xiong2024mqe} and \cite{an2024solving}, we maintain $r^m_{\text{task}}$ and $r^m_{\text{penalty}}$ mentioned in Sec. \ref{subsubsec:mid-level reward}. In addition, extra heuristic rewards, $r^m_{\text{approach}}$ and $r^m_{\text{vel}}$, are added to promote the long-horizon pushing performance of the mid-level network. 

\textbf{High-Level+Low-Level with Fine-Tuned Rewards (H+L FT)} uses the same policy architecture as H+L, incorporating our fine-tuned reward functions. Specifically, We retain $r^h_{\text{task}}$ and $r^h_{\text{penalty}}$, while introducing heuristic rewards similar to $r^m_{\text{heuristic}}$ to this high-level subgoal-proposing policy. This baseline further improves the method from \cite{nachum2019multi} and provides an ablation of our approach, excluding the RRT planner and the mid-level pushing policy.

\textbf{Mid-Level+Low-Level with Fine-Tuned Rewards (M+L FT)} follows the same policy architecture as M+L with our fine-tuned reward functions.
We maintain a complete set of rewards, $r^m_{\text{task}}$, $r^m_{\text{penalty}}$, and $r^m_{\text{heuristic}}$ mentioned in Sec. \ref{subsubsec:mid-level reward}. This baseline further enhances the methods in \cite{xiong2024mqe} and \cite{an2024solving}, while also offering an ablation study of our approach without the high-level controller.

\subsection{Simulation Results and Analysis}
\subsubsection{Comparisons with Baselines}\label{sec:main_result}
Training was carried out in 500 environments, accumulating 80 million steps, each 10 million steps taking approximately one hour of simulation time. The performance of each method is summarized in Table \ref{tab:main_result}. The success rate (S.R.) is evaluated over 50 trials, averaging results from four random seeds with a frozen low-level policy. The completion time (C.T.) represents the average time taken by the robots to complete the pushing task in 50 trials. If a failure occurs during the task, the completion time is recorded as the timeout duration.


As shown in Table \ref{tab:main_result}, the proposed method achieves an average success rate exceeding 60\% in all tasks involving three distinct objects. In contrast, the \textbf{Single-Agent} method exhibits a success rate below 25\%, as a single robot has limited strength and manipulation range for efficient and adaptive pushing of a large and heavy object.
The \textbf{H+L} method also faces challenges, with a similarly low success rate. In some situations, the object movements towards the final goal are disrupted by the other one or two agents, even if the agents can usually reach their own subgoals.
This highlights the difficulty of aligning agent-wise subgoals with effective push control, particularly given the complexity of object interactions, indicating the need for a more fine-grained coordination. 
The \textbf{M+L} method outperforms \textbf{H+L} by enabling more elaborate collaboration, but it exhibits greater variability across different seeds. 
Likewise, it still struggles with the long-horizon nature of the task, as its higher-level policy has difficulty effectively guiding the object towards the distant target.
After training with our designed reward functions, both the \textbf{H+L FT} and \textbf{M+L FT} methods exhibit notable performance improvements, demonstrating the effectiveness of our heuristic rewards, including the OCB reward. However, the \textbf{M+L FT} method still falls significantly short compared to our approach. Additionally, while the \textbf{H+L FT} method achieves a reasonable average performance, it lacks stability and shows poor adaptability with certain objects, such as the T-shape. This indicates the necessity of our designed hierarchical architecture.

\subsubsection{Ablation Study}
\paragraph{The OCB Reward}
To assess the effectiveness of the OCB reward in training the mid-level controller for short-horzion pushing, we conduct an ablation study in a free space environment, where a 6$\,$kg cuboid (\(1.5\,\text{m} \times 1.5\,\text{m} \times 0.5\,\text{m})\) is placed with random orientations. Two agents are randomly initialized, while the target object position (subgoal for the mid-level controller) is generated within a circular area 1.5$\,$m to 3.0$\,$m from the cuboid's initial position randomly. 
As shown in Table \ref{tab:ocb_ablation}, our method significantly outperforms the ablation experiment in both success rate (S.R.) and completion time (C.T.). In particular, as the duration of the timeout increases, the success rate of our method improves more rapidly, indicating a better adaptability to adjust the direction of pushing when the object deviates, an issue that often causes failures with shorter timeouts.

\begin{table}[htp]
\centering
\caption{The results of the OCB-reward ablation study. Two methods are evaluated under two timeout conditions. }
\label{tab:ocb_ablation}
\begin{tabular}{cccccc}                         
\toprule
                                                              & \multicolumn{2}{c}{Timeout=20s} & \multicolumn{2}{c}{Timeout=40s} \\
\cmidrule(lr){2-3} \cmidrule(lr){4-5}
\multicolumn{1}{l}{}                                          & S.R.$\uparrow$         & C.T.$\downarrow$            & S.R.$\uparrow$          & C.T.$\downarrow$    \\         
\midrule
\textbf{Ours}                                                 & \textbf{57.0$\pm$6.1\%}        & \textbf{14.9$\pm$0.8}        & \textbf{74.0$\pm$6.9\%}        & \textbf{22.5$\pm$2.4}     \\   
Ours w/o OCB  & 13.0$\pm$1.2\%                                        & 18.3$\pm$0.2        & 19.0$\pm$2.6\%        & 34.9$\pm$0.3        \\
\bottomrule
\end{tabular}
\end{table}

\begin{figure}[H]
    \centering
    \begin{subfigure}[b]{0.23\textwidth}
        \centering
        \includegraphics[width=\textwidth]{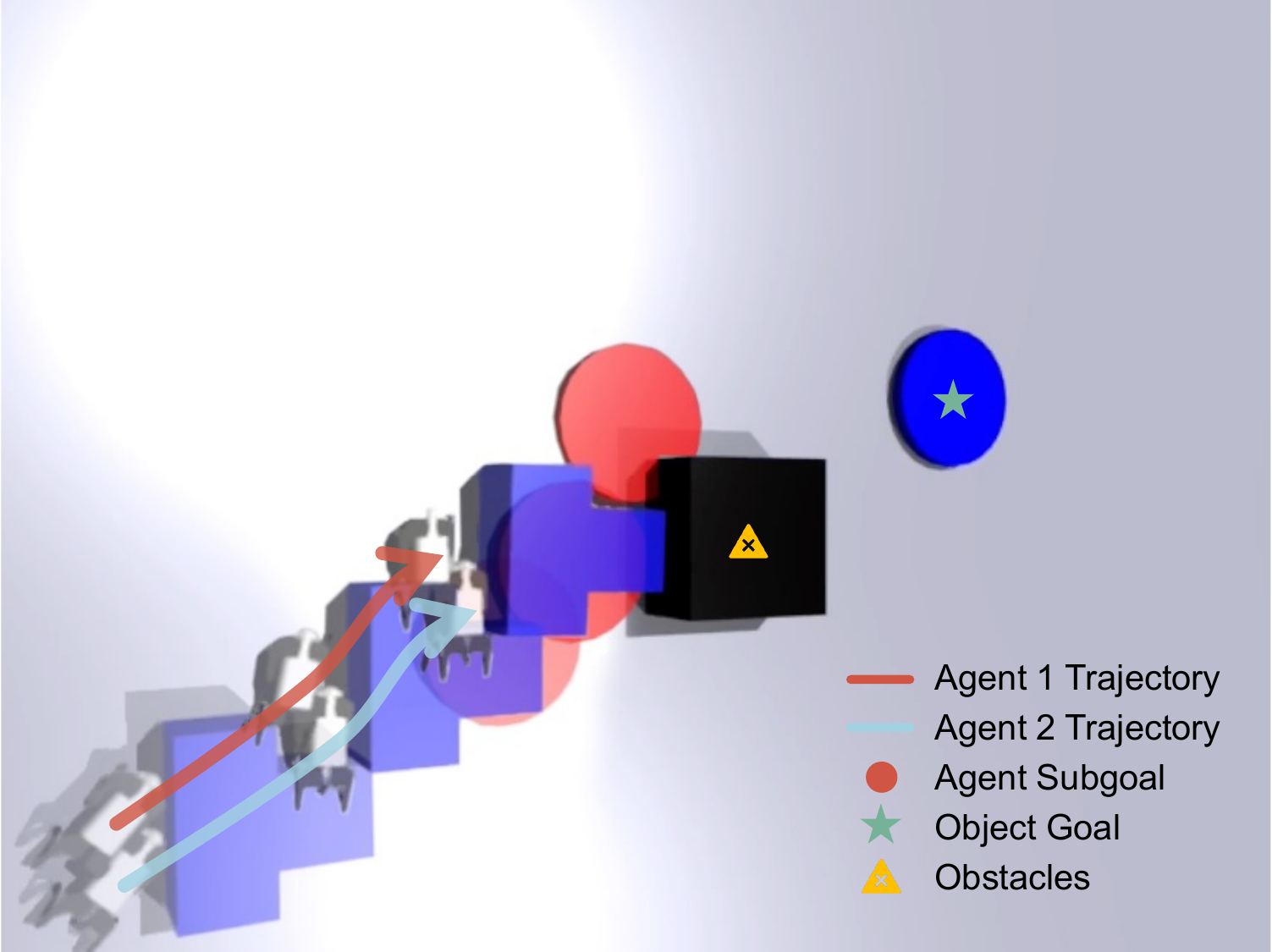}
        \caption{Ours w/o adaptive policy.}
        \label{subfig:adaptive/RRT}
    \end{subfigure}
    \hfill
    \begin{subfigure}[b]{0.23\textwidth}
        \centering
        \includegraphics[width=\textwidth]{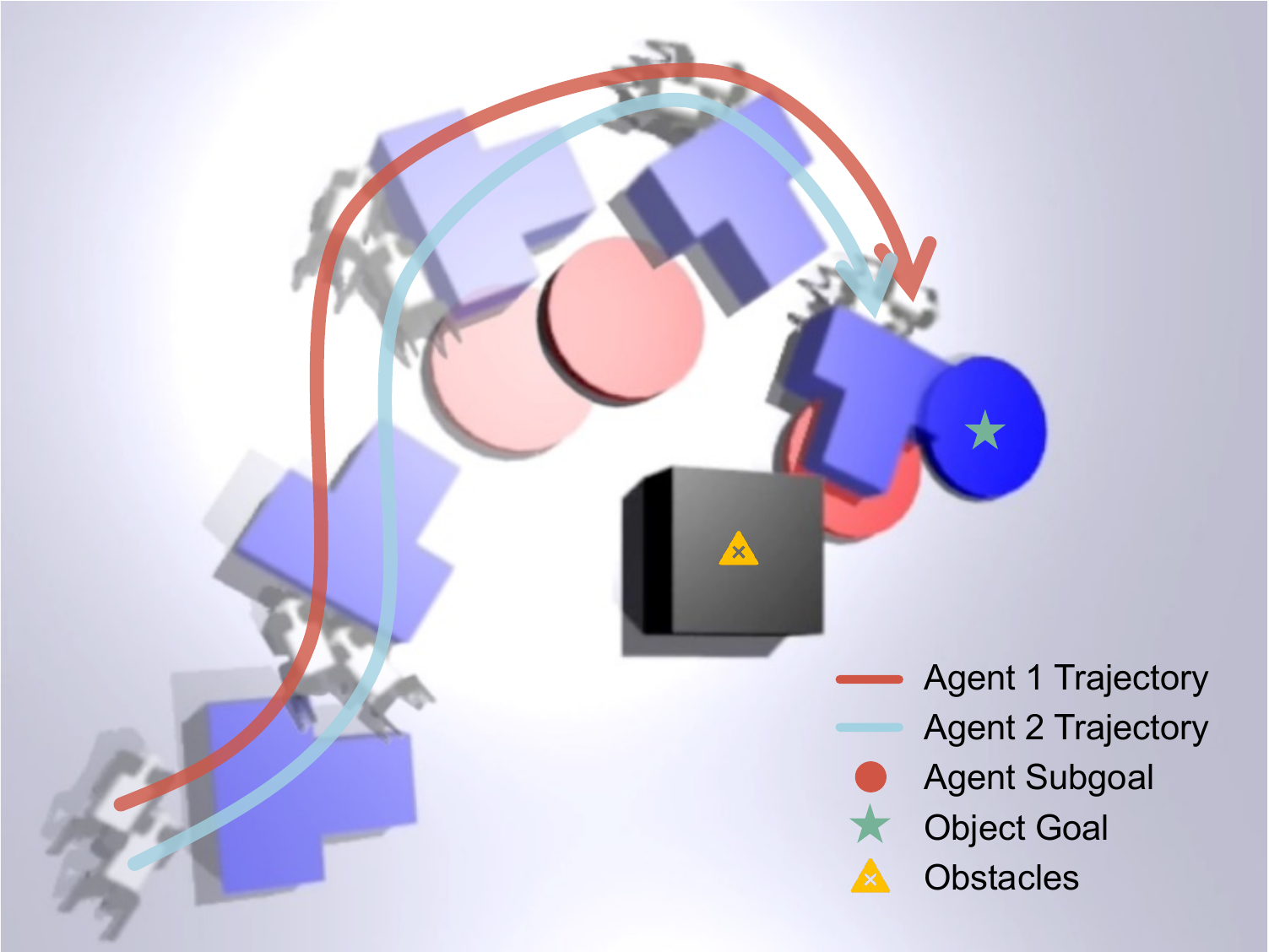}
        \caption{Ours.}
        \label{subfig:adaptive/ours}
    \end{subfigure}
    \label{fig:adaptive}
    \caption{Comparison between our method and the one with only the RRT planner at the high-level controller.}
\end{figure}

\paragraph{The High-Level Adaptive Policy}
To demonstrate the need for an adaptive high-level controller, we design a challenging scenario with obstacles placed directly between the start and target positions of the T-block. Although the RRT planning algorithm finds a short path, it often leads to trajectories that come too close to obstacles without considering the object shape, failing to account for the dynamics of the pushing process and lacking real-time adjustments based on the object's state (Fig. \ref{subfig:adaptive/RRT})
In contrast, our method deviates from obstacles in advance, allowing the robots to bypass them and reach the goal (Fig. \ref{subfig:adaptive/ours}), underscoring the importance of the RL-trained high-level adaptive policy.

\begin{figure}[H]
    \centering
    \includegraphics[width=0.92\linewidth]{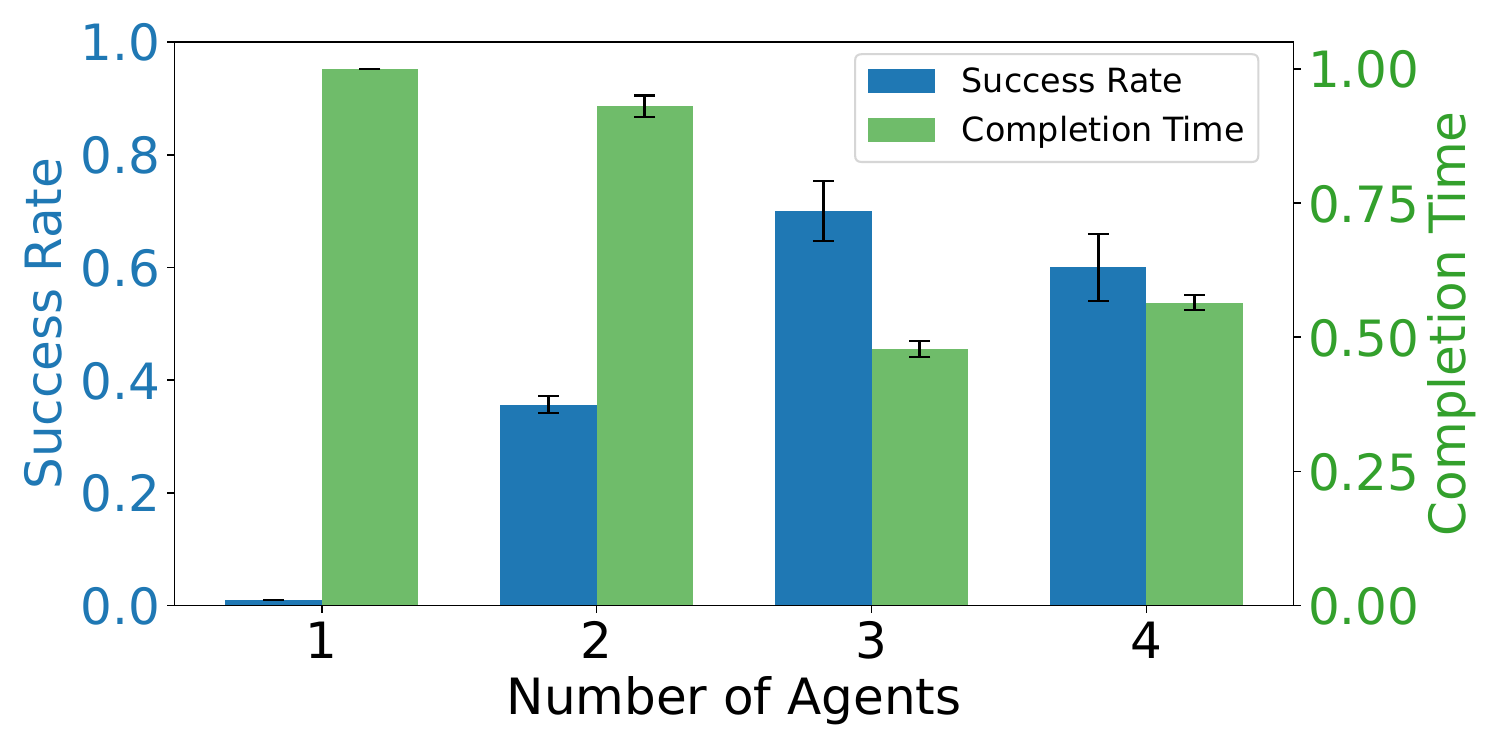}
    \caption{The success rate and completion time of different numbers of robots in the task of cylinder pushing.}
    \label{fig:scalibility_result}
\end{figure}

\begin{figure*}[h]
    \centering
    \includegraphics[width=0.98\textwidth]{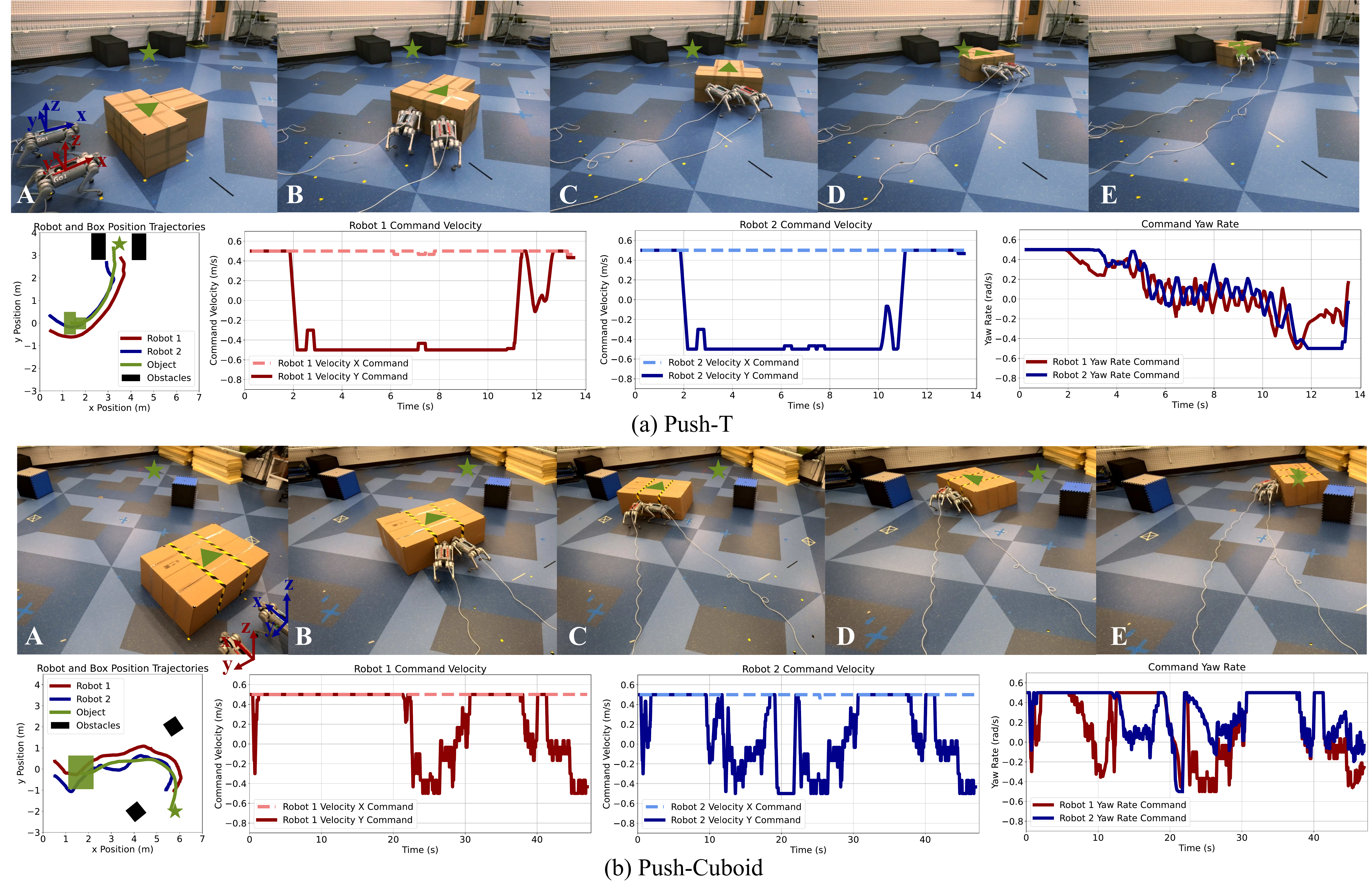} 
    \caption{The results of the physical-robot experiments where \textbf{\textcolor{BrickRed}{Robot 1}} and \textbf{\textcolor{BlueViolet}{Robot 2}} collaboratively push an \textbf{\textcolor{OliveGreen}{Object}} to its \textbf{\textcolor{OliveGreen}{Target Position}}, while avoiding \textbf{Obstacles}. The first row of each section demonstrates video snapshots corresponding to the each task completion, with the local frames of \textbf{\textcolor{BrickRed}{Robot 1}} and \textbf{\textcolor{BlueViolet}{Robot 2}} indicating their respective velocity command references. The leftmost figure of each second row illustrates the motion trajectories of both robots and the \textbf{\textcolor{OliveGreen}{Object}}. The middle two figures of each second row depict the linear velocity commands received by \textbf{\textcolor{BrickRed}{Robot 1}} and \textbf{\textcolor{BlueViolet}{Robot 2}}, respectively. The rightmost figure of each second row displays the angular velocity commands received by both robots. The three figures on the right of each task are all processed with a sliding window of size 15.}
    \label{fig:real_world_experiment}
\end{figure*}


\subsubsection{Scalability Analysis}
To evaluate the scalability of the proposed method, we experiment with different numbers of robots to push the cylinder described in Sec. \ref{subsubsec:env and task}. 
As shown in Fig. \ref{fig:scalibility_result}, a single robot struggles with cylinder pushing, while two robots significantly improve success rate with reasonable completion time. Three robots further enhance both metrics, achieving the best performance. However, with four robots, performance slightly declines due to the cylinder's limited effective contact surface, increasing collision risk and pushing interference (see attached video).


\subsection{Real-World Setups}
\subsubsection{Environments and Task}
We evaluate our policy on real hardware in a 7.5$\,$m $\times$ 7.5$\,$m room, utilizing two Unitree Go1 robots, each with an approximate payload of 5$\,$kg. The entire room is equipped with 24 Prime$^\text{X}$ 22 cameras, using OptiTrack's motion capture system to gather real-time data regarding robots and objects. We deploy both our high-level and mid-level policies, trained in simulation, on physical robots, utilizing the integrated locomotion policy of the Unitree Go1 as the low-level controller for real-world experiments. 


We conduct two tasks for two robots: pushing a cuboid block and pushing a T-shaped block. The cuboid utilized in the experiment measures $1.5\,\text{m} \times 1.0\,\text{m} \times 0.5\,\text{m}$ and weighs 6.8$\,$kg. The T-block consists of a main body measuring $0.5\,\text{m} \times 1.0\,\text{m} \times 0.5\,\text{m}$, with a protruding section measuring $0.5\,\text{m} \times 0.5\,\text{m} \times 0.5\,\text{m}$, and has an overall weight of 3.3$\,$kg. 
Specifically, in the cuboid task, the target position is randomly set on the opposite side of the room, constrained within an $x$-range of 5.5$\,$m to 6.5$\,$m and a $y$-range of -3.5$\,$m to 3.5$\,$m. In the T-pushing task, the target position is set within an $x$-range of 3.5$\,$m to 4.5$\,$m and a $y$-range of -4$\,$m to 4$\,$m. Even with a limited size of the physical environment, this setup ensures that our pushing process remain sufficiently long to meet the requirements of long-horizon tasks. Additionally, in the cuboid-pushing task, obstacles are randomly initialized within a narrow 2$\,$m band surrounding the line connecting the starting and target positions of the box.

\subsection{Real-World Results and Analysis}

\subsubsection{Pushing T}
As shown in Figure~\ref{fig:real_world_experiment}(a), the robots effectively control the turning of the T-shaped block, and successfully push the T-block to the target position. Due to the smaller size of the T-shaped block, finding appropriate pushing points is crucial for the task. We observe that the two robots consistently push from the two ends of the block to apply a larger contact surface, which ensures the application of continuous forward force while also maintaining directional control. 
Additionally, we find that the $x$ direction velocity commands output by the mid-level policy almost always reach the upper bound of 0.5$\,$m$/$s. This outcome is expected because the robot is trained to finish the task within shorter time. Once the robot identifies the correct pushing points, its motion becomes predominantly forward, with turning primarily achieved through adjustments in its yaw. Since the target is located in the positive $y$ direction, we observe that the yaw command is initially positive and gradually returns to zero or negative after the turn is completed to straighten the robots. This process is complemented by minor adjustment on the $y$ direction velocity commands, which helps maintain good contact positions for both robots when interacting with the smaller T-shaped object.

\subsubsection{Pushing Cuboid}

The cuboid-pushing task exemplifies a typical multi-robot collaboration challenge, primarily due to the cuboid's large size and weight. It involves long-distance pushing with obstacles, testing the planning and coordination capabilities of our hierarchical framework. As shown in Figure~\ref{fig:real_world_experiment}(b), the robots navigate along an adaptive trajectory, avoiding obstacles and reaching the target. The cuboid's large surface provides ample contact points, enabling the robots to adopt varied pushing postures as needed. Throughout the task, we observe the robots adopting different pushing strategies: at times, both robots push forward with their heads to maximize speed, while at other moments, one or both shift to a sideways push to facilitate turning. The $x$-direction velocity commands frequently reach the upper limit of 0.5$\,$m$/$s, similar to the push-T task, while $y$-direction and yaw rate commands adapt accordingly to refine the pushing strategies.

\subsubsection{Analysis on the Action Homogeneity}


Figure \ref{fig:real_world_experiment} reveals similar motion patterns and a degree of homogeneity in the $xy$ commands. This behavior can be attributed to several factors: (1) \textbf{\textcolor{BrickRed}{Robot 1}} and \textbf{\textcolor{BlueViolet}{Robot 2}} are treated as homogeneous agents and utilize a shared mid-level pushing policy network; (2) both robots must maintain contact with the same surface of the object for efficient pushing, resulting in similar states and observations; and (3) the yaw rate commands primarily handle strategic adjustments and motion coordination, allowing the $xy$ commands to remain relatively homogeneous.

\section{Conclusions}
\label{sec:conclusions}
In this paper, we address the challenge of obstacle-aware, long-horizon object pushing by coordinating the whole-body motions of multiple quadrupedal robots. While previous studies make significant progress in improving quadrupedal mobility and some aspects of manipulation, their ability to handle large objects in complex, real-world environments remains limited. To overcome these limitations, we propose a hierarchical MARL framework, consisting of high-level, mid-level, and low-level controllers, to enable effective and coordinated pushing tasks. Through extensive simulation experiments, we demonstrate that our approach significantly outperforms the best baseline methods, achieving a $36.0\%$ higher success rate and a $24.5\%$ reduction in completion time. Through physical robot experiments, our method is validated to effectively handle obstacle-aware, long-horizon tasks such as Push-Cuboid and Push-T, highlighting its potential for real-world applications.





\section*{ACKNOWLEDGMENT}
We would like to express our gratitude to Mengdi Xu for her initial idea brainstorming, early exploration, valuable discussions on the project, and feedback on our manuscript. We also thank Changyi Lin for his efforts in testing the low-level controller in real-robot experiments. We are grateful to Yuyou Zhang for her insightful discussions, constructive feedback, and contributions to figure creation in our paper. Additionally, we are thankful to Zili Tang for his valuable input on the reward design for the mid-level controller.


\bibliographystyle{IEEEtran}
\bibliography{references}

\clearpage
\section*{APPENDIX}
\label{sec:appendix}
	\subsection{Training Details}
The training and setup of the low-level locomotion policy in simulation follow the approach described in \cite{margolis2023walk}. For detailed implementation information, we direct readers to this prior work.

The mid-level controller is to enable the robot team to complete shorter-horizon pushing tasks, specifically those where the distance between the initial position and the target position of the object is less than 3.0$\,$m. The environments for the mid-level policies are designed to be free of obstacles, with a dimension of $24\,$m $\times$ $24\,$m. 
The initial object positions are static, while both the target object positions and the initial agent positions are randomized within a circular area measured by a polar coordinate system represented by $r$ and $\theta$. Additionally, the yaw of both the initial objects and agents is randomized from 0 to $2\pi$. This ensures adequate coverage of partial observation settings. In practice, we find that a substantially high threshold for subgoal reaching will improve the success rate of long-horizon pushing. This prevents the policy from becoming trapped into fine-grained manipulation around an intermediate subgoal, which is intended to serve as guidance toward the final target. 
An episode concludes either when the agent team successfully completes the pushing task or when an exception occurs, including robot fall-overs, collisions, object tilting, or timeouts.
The details of the above environment settings are specialized in Table \ref{tab:env}.


\begin{table}[htp]
\centering
\caption{Environment setups to train the mid-level controller}
\begin{tabular}{cccc}
\toprule
Object  & Cuboid & T-Shaped &  Cylinder  \\
\midrule
\multicolumn{1}{c}{initial agent $r$ range (m)}& $[1.2,1.3]$ & $[1.2,1.3]$  & $[2.0,2.5]$  \\
\multicolumn{1}{c}{initial agent $\theta$ range (rad)}&  $[0, 2\pi]$ & $[0, 2\pi]$  & $[0, 2\pi]$ \\
\multicolumn{1}{c}{initial target $r$ range (m)}& [1.5,3.0] & $[1.5,3.0]$   & $[1.5,3.0]$  \\
\multicolumn{1}{c}{initial target $\theta$ range (rad)}& $[0, 2\pi]$ & $[0, 2\pi]$  & $[0, 2\pi]$ \\
timeout (s)  & 20 & 20 & 20 \\
subgoal reaching threshold (m) & 1.0 & 1.0 & 1.0 \\
\bottomrule
\end{tabular}
\label{tab:env}
\end{table}

In the high-level controller, the RRT planner is utilized to generate a reference path for guidance of the robots during test time. Then, the adaptive policy will produce subgoals for the mid-level controller based on the reference path, environment dynamics, and agent states. Due to the overhead of implementation of RRT planner for massively parallel environments in IssacGym during adaptive policy training, we randomly generate a long curved trajectory to replace the RRT-planned reference path as the input of the adaptive policy, with the start and end points consistent with the object positions and targets of our task settings. Then, we generate obstacles randomly around the trajectory. Specifically, the obstacles are placed within a 4-meter-wide strip area centered around the reference path. This ensures efficient obstacle sampling to train the adjustment capability of the adaptive policy.


\subsection{Reward Details}
The reward functions utilized to train the low-level locomotion policy is aligned with \cite{margolis2023walk}. We direct readers to this prior work for details.

The reward terms to train the mid-level adaptive policy is shown in Table \ref{tab:mid_reward}. The \textcolor{BlueViolet}{task reward $r_{\text{task}}^{m}$} includes the reward for the object to approach or reach the subgoal, where $d_{\text{subgoal},t}$ represents the 2D distance between the object's current position and the subgoal's position at step $t$, and $\alpha=200$ is a coefficient for the delta distance. The \textcolor{BrickRed}{penalty term $r_{\text{penalty}}^{m}$} penalizes exceptions and robot close proximities, where $N$ is the agent number and $d_{i,j}$ is the current distance agent $i$ and $j$. The \textcolor{OliveGreen}{heuristic reward $r_{\text{heuristic}}^m$} is composed of reward for approaching the object, object velocity, and the OCB reward, where $d_{\text{object},i}$ denotes the current distance between the object and agent $i$ and $v_{\text{object}}$ is the velocity of the object.



    
    
    

\begin{table}[htp]
    \centering
    \caption{Reward terms to train the mid-level controller}
    \begin{tabular}{c c c}
        \toprule
        \textbf{Reward}                  & \textbf{Expression}                                             & \textbf{Weight} \\
        \midrule
        \textcolor{BlueViolet}{Subgoal Reaching}                  & $\mathbbm{1} \text{(reach\;subgoal)}$                      & 10          \\
        \textcolor{BlueViolet}{Subgoal Approaching}               & $\alpha(d_{\text{subgoal},t-1} - d_{\text{subgoal},t}) - d_{\text{subgoal},t}$ & 3.25e-3  \\
        \textcolor{BrickRed}{Exception Avoidance}                        & $\mathbbm{1} \text{(exception)}$                             & -5         \\
        \textcolor{BrickRed}{Collision Avoidance}                  & $\sum_{\substack{j \neq i}}^{N} \frac{1}{0.02 + d_{i,j}/3}$   & -2.5e-3           \\
        \textcolor{OliveGreen}{Object Approaching}               & $-(d_{\text{object},i} + 0.5)^2$                             & 7.5e-4             \\
        \textcolor{OliveGreen}{Object Velocity}                  & $\mathbbm{1} (v_{\text{object}} > 0.1)$                       & 1.5e-3             \\
        \textcolor{OliveGreen}{OCB}                              & $\vec{v}_i\cdot\vec{v}_{\text{target}}$ & 4e-3 \\
        \bottomrule
    \end{tabular}
    \label{tab:mid_reward}
\end{table}

The reward terms to train the high-level adaptive policy is shown in Table \ref{tab:high_reward}. 
The \textcolor{BlueViolet}{high-level task reward $r_{\text{task}}^h$} consists of three components: a reward for the object moving toward the final target, a reward for the object reaching the final target, and a reward for the subgoal following the planned trajectory.  Here, $d_{\text{target}}$ represents the Euclidean distance between the object's current position and the target position, while $d^h_{\text{subgoal, path}}$ is the distance between the output subgoal and the nearest sampled point on the reference path. The \textcolor{BrickRed}{penalty term $r^h_{\text{penalty}}$} account for exceptions and penalize situations where the object is in close proximity to obstacles. Here, $d_{\text{obstacle}}$ is the distance between the object and its nearest obstacle.





\begin{table}[htp]
\centering
\caption{Reward terms to train the high-level controller}
\begin{tabular}{ccc}
\toprule
Reward  & Expression & Weight \\
\midrule
\textcolor{BlueViolet}{Target Reaching} & $\mathbbm{1} \text{(reach\;final\;target)}$ &  2  \\
\textcolor{BlueViolet}{Target Approaching}  & $\frac{1}{1 + d_{\text{target}}}$ &  0.3 \\
\textcolor{BlueViolet}{Path Following}  & $\frac{1}{1 + d_{\text{subgoal, path}}}$ &  0.5 \\
\textcolor{BrickRed}{Exception Avoidance}  & $\mathbbm{1} \text{(exception)}$ &  -0.5 \\
\textcolor{BrickRed}{Obstacle Avoidance} & $\frac{1}{1 + d_{\text{obstacle}}}$ & -0.1   \\
\bottomrule
\end{tabular}
\label{tab:high_reward}
\end{table}

\subsection{Hyperparameters}
The hyperparameters of the low-level locomotion policy, the mid-level decentralized pushing policy, and the high-level adaptive policy are detailed in Table \ref{tab:low_hyper}, Table \ref{tab:mid_hyper}, and Table \ref{tab:high_hyper}, respectively. 

\begin{table*}[htp]
\centering
\caption{Hyperparameters of PPO for the low-level controller training.}
\begin{tabular}{cc}
\toprule
Hyperparameter  & Value \\
\midrule
number of environments  &  4096 \\
leaning rate  &  1e-3 \\
discount factor  &  0.99 \\
gae lambda  &  0.95 \\
batch size & 4096 $\times$ 24 \\
number of epochs  & 5 \\
number of minibatches per epoch  & 4 \\
value loss coefficient & 1.0 \\
clip range & 0.2 \\
entropy coefficient & 0.01 \\
optimizer  &  Adam \\
\bottomrule
\end{tabular}
\label{tab:low_hyper}
\end{table*}

\begin{table*}[htp]
\centering
\caption{Hyperparameters of MAPPO for the mid-level decentralized policy training. $N$ is the number of agents.}
\begin{tabular}{cc}
\toprule
Hyperparameter  & Value \\
\midrule
number of environments  & 500 \\
leaning rate  &  5e-4 \\
discount factor  & 0.99 \\
gae lambda  &  0.95 \\
number of epochs & 10\\
batch size  & $500 \times 200 \times N$  \\
value loss coefficient &  0.5\\
clip range & 0.2 \\
entropy coefficient & 0.01\\
optimizer  &  Adam \\
\bottomrule
\end{tabular}
\label{tab:mid_hyper}
\end{table*}

\begin{table*}[htp]
\centering
\caption{Hyperparameters of PPO for the high-level adaptive policy training.}
\begin{tabular}{cc}
\toprule
Hyperparameter  & Value \\
\midrule
number of environments  &  500 \\
leaning rate  &  5e-4 \\
discount factor  &  0.99 \\
gae lambda  &  0.95 \\
number of epochs & 10 \\
batch size  &  500 $\times$ 200 \\
value loss coefficient & 0.5 \\
clip range & 0.2 \\
entropy coefficient & 0.01 \\
optimizer  &  Adam \\
\bottomrule
\end{tabular}
\label{tab:high_hyper}
\end{table*}


\end{document}